%% file: main.tex
\documentclass[10pt,twocolumn,letterpaper]{article}

\usepackage{cvpr}              


\usepackage{amsmath}  
\usepackage[ruled,vlined]{algorithm2e}  
\definecolor{cvprblue}{rgb}{0.21,0.49,0.74}
\usepackage[pagebackref,breaklinks,colorlinks,citecolor=cvprblue]{hyperref}
\usepackage{stfloats}

\newcommand{\blfootnote}[1]{%
  \begingroup
  \renewcommand\thefootnote{}
  \footnotetext{#1}%
  \endgroup
}

\def\paperID{9096} 
\def\confName{CVPR}
\def\confYear{2025}

\title{Towards High-Fidelity 3D Portrait Generation with Rich Details \\by Cross-View Prior-Aware Diffusion}
\author{Haoran Wei$^{1*}$, Wencheng Han$^{1*}$, Xingping Dong$^{2}$, Jianbing Shen$^{1\dagger}$\\
$^{1}$SKL-IOTSC, CIS, University of Macau, $^{2}$School of Computer Science, Wuhan University\\
{\tt\small \{hr.wei1998, wenchenghan, xingping.dong\}@gmail.com, jianbingshen@um.edu.mo}}

\begin{document}
\twocolumn[{%
\renewcommand\twocolumn[1][]{#1}%
\maketitle
\begin{center}
    \centering
    \captionsetup{type=figure}
    \includegraphics[width=0.98\textwidth]{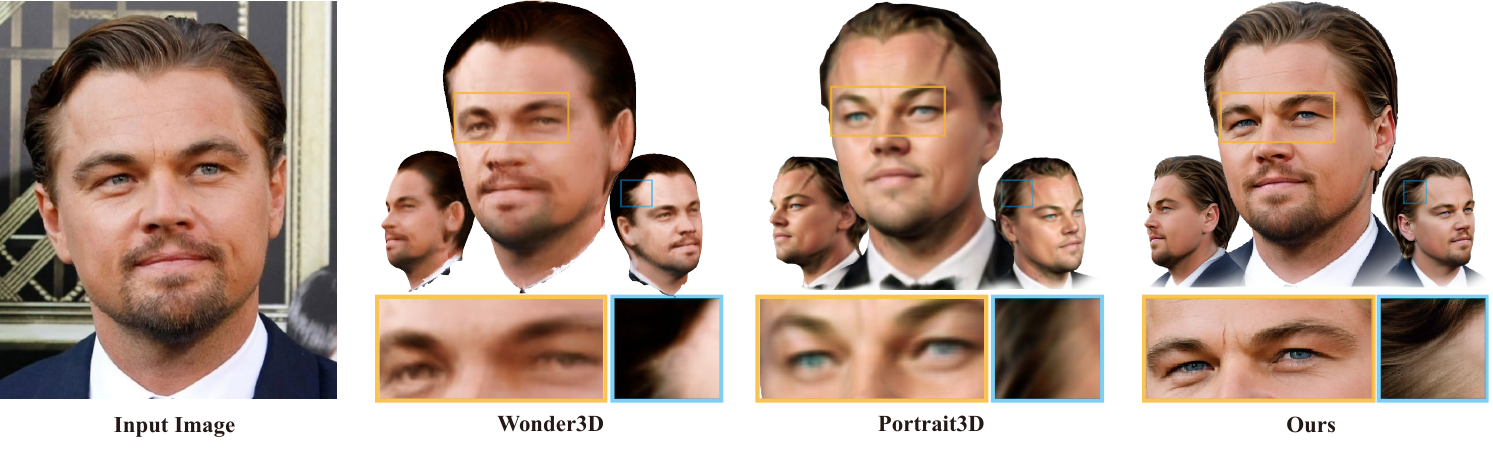}
    \caption{Our proposed \textbf{{Portrait Diffusion}} framework can generate high-quality detail-rich 3D portraits from a single reference portrait image. In comparison to SOTA methods \textbf{Wonder3D}~\cite{long2024wonder3d} and \textbf{Portrait3D}~\cite{wu2024portrait3d}, our approach achieves clearer and more detailed textures. }\label{fig:Nvs_motivation}
\end{center}%
}]

\blfootnote{$*$ Equal contribution. $\dagger$ Corresponding author: \textit{Jianbing Shen}.}

\begin{abstract}
Recent diffusion-based Single-image 3D portrait generation methods typically employ 2D diffusion models to provide multi-view knowledge, which is then distilled into 3D representations.
However, these methods usually struggle to produce high-fidelity 3D models, frequently yielding excessively blurred textures.
We attribute this issue to the insufficient consideration of cross-view consistency during the diffusion process, resulting in significant disparities between different views and ultimately leading to blurred 3D representations. 
In this paper, we address this issue by comprehensively exploiting multi-view priors in both the conditioning and diffusion procedures to produce consistent, detail-rich portraits.    
From the conditioning standpoint, we propose a Hybrid Priors Diffsion model, which explicitly and implicitly incorporates multi-view priors as conditions to enhance the status consistency of the generated multi-view portraits.
From the diffusion perspective, considering the significant impact of the diffusion noise distribution on detailed texture generation, we propose a Multi-View Noise Resamplig Strategy integrated within the optimization process leveraging cross-view priors to enhance representation consistency. 
Extensive experiments demonstrate that our method can produce 3D portraits with accurate geometry and rich details from a single image. The project page is at \url{https://haoran-wei.github.io/Portrait-Diffusion}.
\end{abstract}


\section{Introduction}

The generation of realistic 3D portraits from a single image \cite{deng2024portrait4d, xiang2020one, doukas2021headgan,wu20233dportraitgan,ma2023otavatar} has become an important focus in computer vision and graphics, with broad applications in augmented reality, virtual reality, video conferencing, and gaming\cite{jiang2024mobileportrait,li2024generalizable,ye2024real3d}.
The most straightforward approach involves training GAN models \cite{yin2023gan, an2023panohead} on extensive portrait datasets to directly produce 3D representations. However, acquiring such training data can be costly and technically challenging, leading to failures in generating high-fidelity 360° full-head portraits \cite{deng2024portrait4d, deng2024portrait4dv2} and often resulting in a lack of diversity in the outputs.

To address these limitations, recent developments \cite{lin2022dreamfusion, yi2024gaussiandreamer, qian2023magic123, tang2024makeitvivid, tang2023makeit3d} leverage text-to-image diffusion priors \cite{yang2023uni,xie2023smartbrush,corneanu2024latentpaint}, which exhibit stronger generalization capabilities and higher generation quality, to produce novel perspectives.
Most approaches incorporate additional priors, such as reference image latents \cite{xie2024x, zhang2024rodinhd}, ID features \cite{shao2024human4dit,hao2024portrait3d}, and view embeddings \cite{shao2024human4dit}, to enhance the consistency between new perspectives and the primary viewpoint.
Subsequently, they commonly employ Score Distilling Sampling (SDS) loss \cite{poole2022dreamfusion} to distill these 2D priors into 3D representations, ensuring consistent 3D generation.



However, in single-image 3D portrait generation, these methods still face challenges: generated portraits often appear over-smoothed and fail to capture detailed textures like hair strands, as illustrated in \cref{fig:Nvs_motivation}, limiting their practical applications. 
We attribute this issue to the insufficient consideration of cross-view consistency during the diffusion process, resulting in significant disparities between different views. This 2D inconsistency results in blurred 3D output by SDS optimization.
Although these methods attempt to improve consistency by incorporating additional priors, they rely solely on diffusion attentions to implicitly convey these priors. This reliance results in a lack of explict constraints, leading to inconsistent status across different viewpoints.
Moreover, the diffusion procedure is inherently stochastic; even with the same conditions, a diffusion model can generate varied representations due to  randomly sampled noises. By using view-independent procedures with purely random noise in diffusion, these methods overlook the impact of stochasticity on representation consistency.
Consequently, these inconsistencies in status and representation jointly result in over-smoothed 3D models when optimized under the SDS loss, which enforces 3D consistency and continuity in sacrifice of texture details.

To address these issues, we propose fully exploiting cross-view priors in both the conditioning and diffusion procedures to enhance multi-view consistency, thus yielding detail-rich 3D portraits, as showcased in \cref{fig:Nvs_motivation}. 
From a conditioning perspective, we propose Hybrid Priors Diffusion Model (HPDM). Our approach seeks to transfer and utilize cross-view prior information in both explicit and implicit ways to control the novel view generation.
In an explicit manner, we begin by employing geometric priors to map pixels from the current view to the next, providing an explicit reference to dominate the generation process.
Given that this reference encompasses only a limited overlapping region and contains artifacts introduced through perspective transformations, we further propose to utilize the robust modeling capabilities of attention mechanisms to mitigate these deficiencies. These mechanisms capture finer texture and geometry priors and implicitly transfer these priors into the control conditions, ensuring a more comprehensive and precise guidance for the portrait status of novel viewpoint.
From a diffusion procedure perspective, our goal is to manage randomness in adjacent viewpoints so that they can share detailed, consistent representations. 
To achieve this, we introduce a Multi-View Noise Resampling Strategy (MV-NRS) integrated into the SDS loss, which manages each view's noise distribution by passing cross-view priors.
MV-NRS consists of two main components: first, a shared anchor noise initialization that leverages geometric priors to establish a preliminary representation; 
and second, an anchor noise optimization phase, where we resample and update the anchor noise based on denoising gradient consistency prior to progressively align the representations during the SDS optimization. 

To summarize, our main contributions are as follows:
\begin{itemize}
\item We developed a Portrait Diffusion pipeline consisting of GAN-prior Initialization, Portrait Geometry Restoration, and Multi-view Diffusion Refinement modules to generate rich-detail 3D portraits.

\item We designed a Hybrid Priors Diffusion Model that emphasizes both explicit and implicit integration of multi-view priors to impose conditions, aiming to enhance the consistency of multi-view status.

\item We introduced a Multi-View Noise Resampling Strategy integrated within the SDS loss to manage randomness across different views through the transmission of cross-view priors, thereby achieving fine-grained consistent representations.

\item  Through extensive experiments, we show that our proposed pipeline successfully achieves high-fidelity 3D full portrait generation with rich details.
\end{itemize}




\begin{figure*}[ht]
    \centering
    \includegraphics[width=1.0\textwidth]{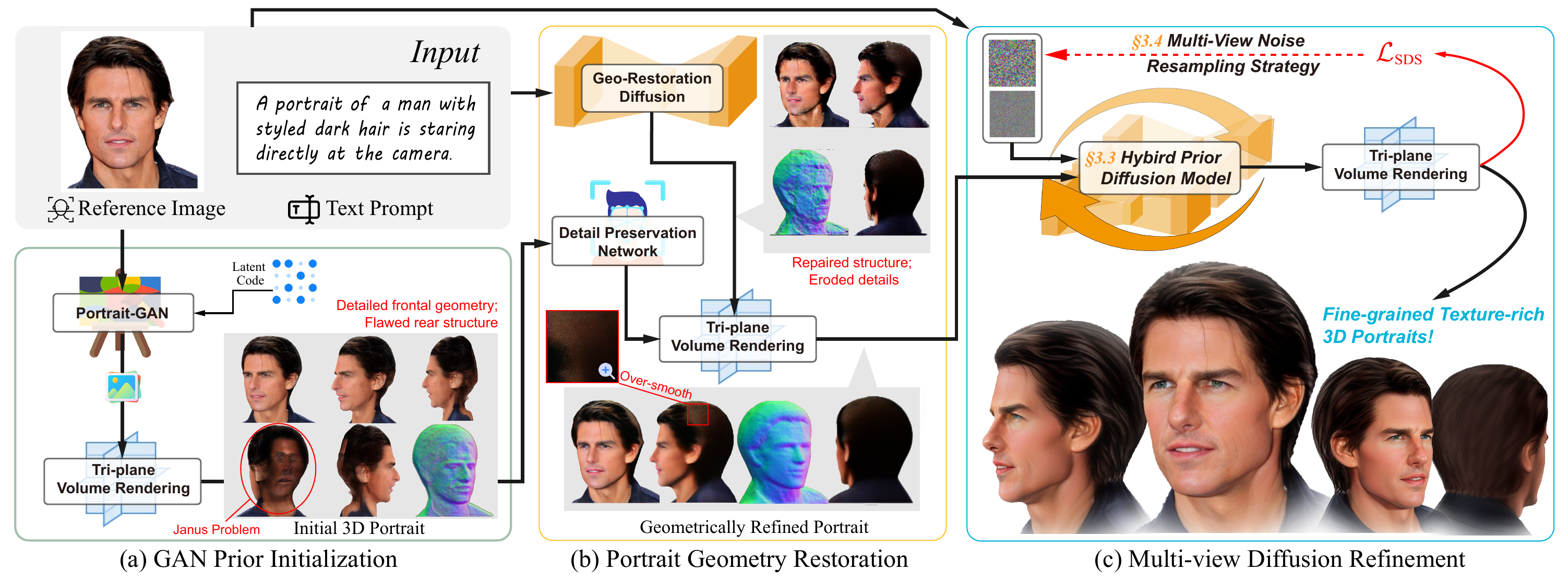} 
    \caption{
 \textbf{The Portrait Diffusion Framework.} This framework comprises three integral modules. 
   \emph{GAN-prior Portrait Initialization}, employs existing Portrait GAN priors to derive initial tri-plane NeRF features from frontal-view portrait images. 
 \emph{Portrait Geometry Restoration}, is focused on reconstructing the geometry using these initialized tri-planes. 
    \emph{Multi-view Diffusion Texture Refinement}, transforms coarse textures into detailed representations.
    }
    \label{fig:pipe}
    \vspace{-4mm}
\end{figure*}

\section{Related Work}
\label{sec:Related Work}

\noindent\textbf{One-shot 3D Generation}
3D GANs \cite{yin20233d, dundar2023progressive, reddy2024g3dr, chen2023single, xiong2023get3dhuman, xie2023high} have made significant strides in advancing one-shot 3D object generation by enhancing both quality and efficiency.  
GRAM \cite{deng2022gram} enhanced efficiency through point sampling on 2D manifolds, and GET3D \cite{gao2022get3d} integrated differentiable rendering with 2D GANs to efficiently generate detailed 3D meshes.
For improving 3D consistency, Geometry-aware 3D GAN \cite{chan2022efficient} used a hybrid architecture to maintain multi-view consistency, while GRAM-HD \cite{xiang2022gramhd} employed super-resolution techniques to address inconsistency issues.
Despite these advances, limited datasets constrain the prior distribution, and acquiring high-quality data remains costly.

Recently, methods leveraging 2D diffusion prior \cite{mirzaei2023spin,mirzaei2023reference,mirzaei2023reference,mirzaei2024reffusion,cao2024mvinpainter,weber2024nerfiller,zeng2024paint3d,wang2024innerf360} for generating 3D objects have gained traction \cite{ding2024text, tewari2023diffusion, karnewar2023holodiffusion, xiang20233d, yi2024gaussiandreamer, qian2023magic123}. 
Dreamfusion \cite{lin2022dreamfusion} introduces a loss mechanism based on probability density distillation for optimizing parametric image generators. 
DreamCraft3D \cite{sun2023dreamcraft3d} employs view-dependent diffusion models for coherent 3D generation, using Bootstrapped Score Distillation to enhance textures. 
Make-It-3D \cite{tang2023makeit3d} uses 2D diffusion models as perceptual supervision in a two-stage process, enhancing textures with reference images. 
Make-it-Vivid \cite{tang2024makeitvivid} focuses on automatic texture generation from text instructions, achieving quality outputs in UV space. 
These advancements underscore the promise of diffusion priors in achieving multi-view consistency in 3D object generation.

\noindent\textbf{One-shot 3D Portrait Generation}
In 3D portrait synthesis, Yin et al. \cite{yin2023gan} enhanced 3D GAN inversion using facial symmetry and depth-guided pseudo labels for better structural consistency and texture fidelity. PanoHead \cite{an2023panohead} creates 360° portraits with a two-stage registration process using tri-mesh neural volumetric representation. 

Benefiting from diffusion priors, diffusion models significantly enhance 3D portrait synthesis by enabling detailed zero-shot full head generation. Portrait3D \cite{wu2024portrait3d} uses 3DPortraitGAN to produce 360° canonical portraits, addressing ``grid-like'' artifacts with a pyramidal tri-grid representation and improving details through diffusion model fractional distillation sampling.
DiffusionAvatars \cite{kirschstein2024diffusionavatars} combine a diffusion-based renderer with a neural head model, using cross-attention for consistent expressions across angles.
Another Portrait3D framework \cite{hao2024portrait3d} by Hao et al. emphasizes identity preservation in avatars across three phases: geometry initialization, sculpting, and texture generation, employing ID-aware techniques.
While many of these methods utilize SDS and incorporate ID and normal information for enhanced representation, they often struggle to fully utilize multiple priors across viewpoints, leading to texture issues like over-smoothing or artifacts.

\section{Methods}
In this section, we first analyze the limitations of existing methods and {give our motivations} (\cref{sec:PF}). Next, we provide an overview of our pipeline, including GAN Prior Initialization Module, Portrait Geometry Restoration Module and Multi-view Diffusion Texture Refinement Module  (\cref{sec:Overall Pipeline}). We then focus on the Multi-view Diffusion Texture Refinement Module, emphasizing both {Multi-view Status Consistency} (\cref{sec:Portrait NVS Diffusion}) and {Multi-view Representation Consistency} (\cref{sec:MV-SDS}) to achieve consistent multi-view generation achieving fine texture fidelity in 3D portrait.
%



\subsection{Preliminary}
\label{sec:PF}

Existing diffusion-based methods for generating 3D objects predominantly utilize Score Distillation Sampling (SDS) loss \cite{poole2022dreamfusion} to distill 2D diffusion priors into 3D representations. This process can be formulated as follows:
\begin{equation}
    \Phi^* =  \underset{\Phi}{\arg\min}(\mathcal{L}_{\text{SDS}}(\Phi; \theta)+\mathcal{L}_{\text{ref}}(\Phi; I^\text{ref}))
\end{equation}
where $\Phi$ denotes the parameters of the 3D model, $\mathcal{L}_{\text{SDS}}(\Phi; \theta)$ represents the SDS loss using a diffusion model paramterized by $\theta$, and $\mathcal{L}_{\text{ref}}(\Phi; I^\text{ref})$ is a loss computed from  reference image $I^\text{ref}$. The SDS loss can be formulated as:
\begin{equation}  \label{eq:sds_loss}
\begin{aligned}  
\nabla_\Phi \mathcal{L}_{\text{SDS}}& =  \mathbb{E}_{t, v, \epsilon} \Big[ w_t \left( \epsilon_\theta(z_{t, v}, t, c) - \epsilon \right)  
\cdot \nabla_\Phi \mathcal{R}_\Phi(v) \Big] \\   
z_{t, v} &= \sqrt{\alpha_t} z_v(\Phi) + \sqrt{1 - \alpha_t} \epsilon,\quad\epsilon \sim \mathcal{N}(0, \textbf{I}),  
\end{aligned}  
\end{equation}  
where \( z_{t, v} \) is a noisy latent representation obtained by combining the image latents \(z_v(\Phi)\), which is rendered from viewpoint \( v \) by \(\Phi\), with random noise \( \epsilon \);
\(\epsilon_\theta(z_{t, v}, t, c)\) is a diffusion UNet model that predicts the noise component at each time step \(t\), conditioned on \(c\). \(w_t\) and \(\alpha_t\) are weitghts, and \(\mathcal{R}\) is rendering function.

From \eqref{eq:sds_loss}, the SDS loss aggregates the denoising gradients from all \(v\) to the 3D model parameters \(\Phi\). When the denosing distributions across viewpoints are inconsistent, the SDS loss will produce over-smoothed representations to minimize the overall loss by averaging conflicting gradients, sacrificing the details of each perspective.
The {denoising function} $\boldsymbol{\epsilon_\theta}$ is influenced by both {the conditions} \(\boldsymbol{c}\) and {the distribution of noise} \(\boldsymbol{\epsilon}\) from each viewpoint, making them essential for the quality of the 3D representation.
%

However, previous methods did not fully leverage multi-view priors to effectively control both conditions \(\boldsymbol c\) and noise \(\boldsymbol \epsilon\). This resulted in inconsistent multi-view denoising, making them impossible to generate detailed 3D textures.
Although some of these methods incorporate additional priors like ID features to enhance the condition, 
they rely solely on implicit priors transfer through embeddings and attention mechanisms within the diffusion, while lacking explicit guidances. Therefore, they are unable to effectively constrain the portrait status across dffierent views.
Additionally, these methods focus merely on enhancing conditions \( \boldsymbol c \)  and overlook the significant influence of \(\boldsymbol \epsilon\) on detailed representations.
As a result, the mutually independent multi-view noise adding procedure leads to a lack of fine-grained alignment in denoising gradients.

\begin{figure*}[ht]
    \centering
    \includegraphics[width=0.98\textwidth]{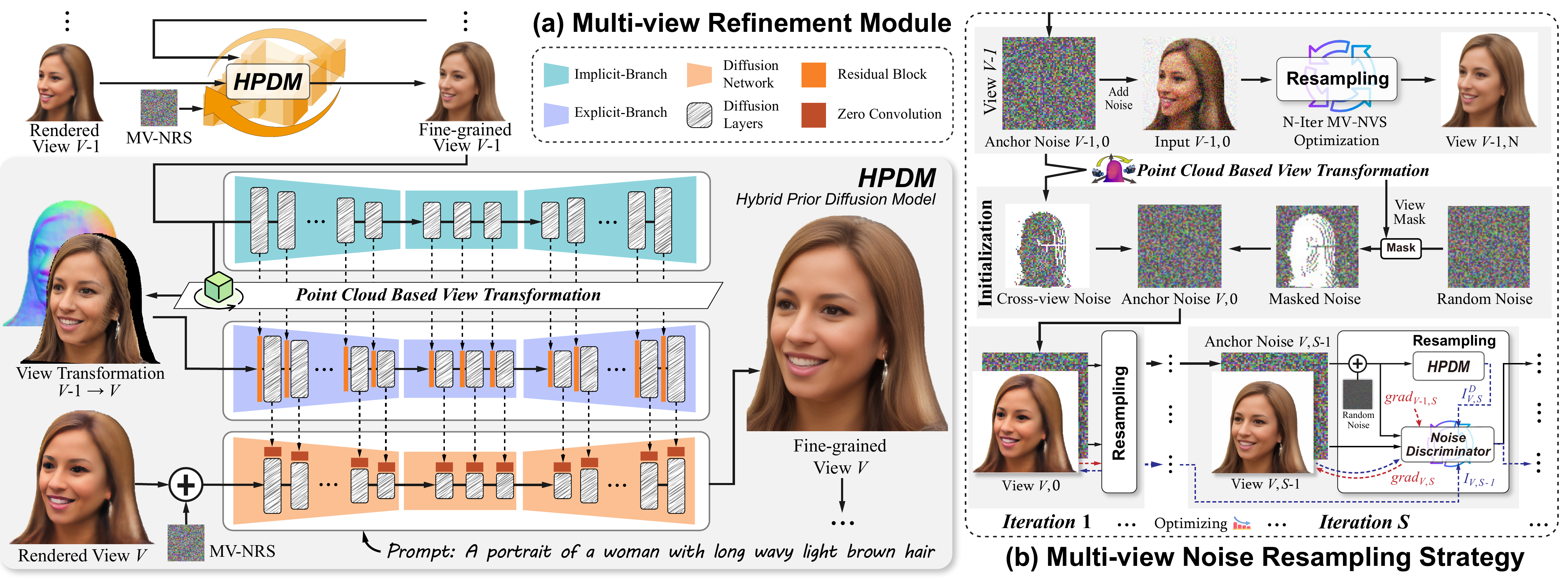} 
    \caption{The presentations of our  proposed \textbf{Hybrid Priors Portrait Diffusion model} (a) and \textbf{Multi-View Noise Resampling Strategy} (b). HPDM is designed to leverage various multi-view priors in a hybrid manner to condition the new view synthetic process for more consistent status. NV-NRS is designed to transfer corss-view priors to control the diffusion noise distribution for representations alignment.}
    \label{fig:p-nvs}
    \vspace{-4mm}
\end{figure*}

\subsection{Detail-Rich Portrait Diffusion}
\label{sec:Overall Pipeline}

Our {Portrait Diffusion} framework for high-fidelity detail-rich 3D portrait generation is illustrated in \cref{fig:pipe}. It consists of three major modules: 
%
\vspace{3pt}



\noindent\textbf{GAN-Prior Initialization Module} utilizes GAN priors learnt from large-scale offline multi-view portrait iamges to initialize a tri-plane representation, as shown in \cref{fig:pipe} (a). This tri-plane offers preliminary geometry and texture, facilitating the subsequent training process. 

In our method, we employ a NeRF, parameterized by $\Psi$, as our 3D model. Initialization plays a critical role in the quality of NeRF models. Standard initialization, such as using a central sphere, often produce excessively smooth geometries with insufficient details. Therefore, a high-quality initialization can significantly benefit the subsequent optimization process.
Inspired by Portrait 3D \cite{hao2024portrait3d}, we utilize GAN priors to initialize NeRF representations.
Specifically, we utilize SOAT GAN method \cite{deng2024portrait4d} to generate triplane features from a frontal image. This process can be formalized as:
\begin{equation}
\begin{aligned}
T_{\text{init},I} &= \mathcal{G}_{\psi_{\text{enc}}}(I) \\ \Psi_{\text{init},I} &= \{ T_{\text{init},I}, \psi_{\text{dec}}, \psi_{\text{SR}} \}
\end{aligned}
\end{equation}
where $G$ denotes the GAN, which is parameterized by the encoder $\psi_{\text{enc}}$, decoder $\psi_{\text{dec}}$ and superresolution $\psi_{\text{SR}}$ parameters, $T$ represents triplane features and $\Psi_{\text{init},I}$ represents the initialized NeRF parameters for image $I$.

While GAN-generated portraits effectively capture frontal details, they are constrained by the lack of 360$^\circ$ priors, leading to missing geometry at the back and the Janus issue.
Therefore, we devise a Portrait Geometry Restoration module to futher repair the geometry for the 3D portrait.


\vspace{3pt}



\noindent\textbf{Portrait Geometric Restoration Module} is designed to fix the structure of the initialized tri-plane with Diffusion priors. It employ diffusion models to deliver high-quality, generalized priors, and introduces a Detail Preservation Block that effectively preserves the details from the initialized priors, as shown in \cref{fig:pipe} (b).
%

Directly optimizing the triplane can lead to the erosion of initialized details; therefore, we employ a Detail Preservation Module that features a small UNet to transform the initialized tri-plane into the desired tri-plane instead. The core idea is to propagate gradients across the entire tri-plane through conv layers, thereby effectively leveraging the global priors. It considers the overall distribution of the initialized priors and subtly adjusts it to maintain coherence through optimizing the UNet paramters.

The Neural Geo Restoration process is divided into two steps: 
First, by pretraining on a large dataset of portraits with SDS loss and Geo-Restoration Diffusion, the UNet, parameterized by $\phi$, acquires generalized structure generation capabilities:
\begin{equation}  
    \phi_\text{pre}^* = \underset{\phi}{\arg\min} \sum_{I \in \mathcal{D}} L_{\text{SDS}}\big(\text{UNet}_{\phi}\left(T_{\text{init},I}\right); \Psi_{\text{init}, I}\big)  ,
\end{equation}
where $\phi$ denotes the parameters of the UNet.

Then, it is fine-tuned on a specific portrait to adapt perfectly to its unique structure:
\begin{equation}  
\begin{aligned}
    \phi_{\text{ft},I}^*, \psi_{\text{dec},I}^* =   
    \underset{\phi, \psi_\text{dec}}{\arg\min} \ L_{\text{SDS}}
    \big(\text{UNet}_{\phi}\left(T_{\text{init},I}\right),\\ \psi_\text{dec}; T_{\text{init},I}, \psi_\text{SR}\big), \quad \text{s.t. } \phi(0) = \phi^*_{\text{pre}} 
\end{aligned}
\end{equation}  
\begin{equation} 
\begin{aligned}
    T^*_{\text{ft},I} &= \text{UNet}_{\phi^*_{\text{ft},I}} (T_{\text{init},I})\\
    \Psi_{0,I} &=\{T_{\text{ft},I}^*, \psi_{\text{dec},I}^*, \psi_\text{SR}\},  
\end{aligned}
\end{equation}  
where $\Psi_{0,I}$ represents the startup NeRF parameters for texture generation of image $I$, \(\phi(0)\) represents the initial value of \(\phi\) for fine-tuning.
More training details are presented in 
Sec. {\color{red} 7}.

Through these two modules, we have achieved geometrically refined portraits; however, intricate texture details are lost due to multi-view inconsistency training. To address this, we have designed a Multi-view Diffusion Refinement Module utilizing this geometric prior.


\vspace{3pt}

\noindent\textbf{Multi-view Diffusion Refinement Module} generates fine-grained 3D texture based on the reconstructed geometry through our Hybrid Priors Diffusion Model and Multi-View Noise Resampling Strategy, as shown in \cref{fig:pipe} (c).

%
This method is designed to thoroughly utilize various priors from both conditioning and diffusion procedure perspectives to improve consistency. 
From a conditioning perspective, the Hybrid Priors Diffusion Model effectively leverages and transmits multi-view priors both explicitly and implicitly—utilizing additional conditioning branches parameterized by $\theta^{\text{Ex}}$ and $\theta^{\text{Im}}$—to enhance the consistency of novel viewpoints. 
From a diffusion procedure perspective, we acknowledge the role of noise in conveying detailed multi-view priors and devised a Multi-View Noise Resampling Strategy integrated within the SDS loss ($\mathcal{L}^\textbf{MV-NRS}_{\text{SDS}}$), which adjusts the distribution of resampled diffusion noise $\epsilon^\text{Rs}$ for fine-grained representations alignment. Through the $\mathcal{L}^\textbf{MV-NRS}_{\text{SDS}}$ and $\theta^{\text{Ex}}, \theta^{\text{Im}}$, we can generate detail-rich portraits:
\begin{equation}
\begin{aligned}
     \Psi^*_{I} =  \underset{\Psi}{\arg\min}&\big(\mathcal{L}^\textbf{MV-NRS}_{\text{SDS}}(\Psi, \epsilon^\text{Rs}; \{\theta, \theta^\textbf{Ex}, \theta^\textbf{Im}\})\\
     &+\mathcal{L}_{\text{ref}}(\Psi; I)\big), \quad \text{s.t.  } \Psi(0) = \Psi_{0,I}
\end{aligned}
\end{equation}

\subsection{Multi-view Status Consistency }
\label{sec:Portrait NVS Diffusion}

\cref{fig:p-nvs} (a) presents our \textbf{H}ybrid \textbf{P}rior \textbf{D}iffusion \textbf{M}odel (\textbf{HPDM}), 
which focuses on leveraging multi-view priors in a hybrid manner to condition the novel view synthesis  process for more consistent portrait status. 
Initially, we leverage {explicit priors} by providing reference images to dominate the generation process, offering direct control and constraints. Following inpainting tasks, we introduce our \emph{Explicit-Branch}. This branch takes the image projected from the driving view to the target view as an explicit reference and extends it to fill in the invisible areas.


To generate this reference, we convert the driven view image \( I_{v_i} \) into a colored 3D point cloud \( P_{v_i} \) using the NeRF-rendered depth map \( D_{v_i} \). Then, a reference image of target view is rendered from this colored point cloud:
\begin{equation}
    I^\text{Proj}_{v_{i+1}} =  \mathcal{R}_{P_{v_i}}(v_{i+1}, I_{v_i})
\end{equation}
Besides, the segmentation mask \( S_{v_i} \) is similarly rendered onto the target view as an auxiliary mask condition \( S^\text{Proj}_{v_{i+1}} \). The \( z^\text{Proj}_{v_{i+1}} \), obtained by encoding \( I^\text{Proj}_{v_{i+1}} \) from an VAE, along with the \( S^\text{Proj}_{v_{i+1}} \) are fed into the diffusion UNet.


The diffusion UNet, following the inpainting method \cite{ju2024brushnet},  is a adapted  version of a pretrained diffusion UNet.  In this adaptation, the cross-attention components are removed to to focus entirely on the reference. Features from this UNet are injected into the frozen layers of the original diffusion UNet layer by layer with zero convs, allowing for dense, pixel-level control over the generation process:
\begin{equation}\label{eq:ex}
\begin{aligned}
    \epsilon_\theta(z_{t},\ t &,\ y)_l = 
    \epsilon_\theta(z_{t},\ t ,\ y)_l\\ &+ w^\text{Ex} \cdot \mathcal{Z}(\epsilon_\theta^{\text{Ex}} ([z^\text{Proj},\ S^\text{Proj},\ z_{t}],\ t)_l), 
\end{aligned}
\end{equation}
where $l$ denotes the $l$ layer of the UNet, $\mathcal{Z}$ denotes zero conv and $w^\text{Ex}$ denotes control weight.

\vspace{4pt}

However, since the reference cannot guide all areas and the degraded priors during the view transformation, relying solely on such explicit priors transfer would introduce noises into control signals.

To address this, we aim to {implicitly leverage priors} to compensate for these deficiencies.
To enhance texture priors, we integrated a second branch, \textit{Implicit-Branch}, for lossless texture understanding, and designed a res-block to semi-explicitly pass this understanding to the Explicit-Branch. In detail, this branch is a copy of a Explicit-Branch that directly takes the driving image $I_{v_{i-1}}$ as input. To ensure effective transfer of these priors to the Explict-Branch, we first explicitly rendering Implicit-Branch latents into the target view and then implicitly integrating them into the Explicit-Branch through res-blocks with zero-convs. We opt for simple res-blocks rather than complex transformers, benefiting from the spatial prior alignment provided by explicit geometric projection. The design of the res-blocks is detailed in the 
Sec. {\color{red} 6}
. This process can be expressed as:
\begin{equation}  \label{eq:im}
  \begin{aligned}  
    \epsilon&_\theta^{\text{Ex}} ([z^\text{Proj}_{v_i},\ S^\text{Proj}_{v_i},\ z_{t,v_i}],\ t)_l \\   
    &= \text{Res}\big[\epsilon_\theta^{\text{Ex}} ([z^\text{Proj}_{v_i},\ S^\text{Proj}_{v_i},\ z_{t,v_i}],\ t)_l, \\&\quad \mathcal{R}_{P_{v_{i-1}}}(v_i, \epsilon_\theta^{\text{Im}} ([z_{v_{i-1}},\ z_{t,v_i}],\ t)_l)\big]
  \end{aligned}  
\end{equation}

Additionally, to compensate for the geometric artifacts in the explicit reference, we incorporate the geometric prior of the current view. To ensure that the generation aligns with the current geometry, the rendered coarse image $I^{R}$ and normal map $N^{R}$  are included as additional conditions within the Explicit-Branch. To enhance the geometry without overshadowing the reference texture, we once again employ a res-block to implicitly merge these conditions with the reference in latent space:
\begin{equation} \label{eq:geo}
    z^\text{Proj} = \text{Res}(z^\text{Proj}, \text{VAE}(N^{R},I^{R}))
\end{equation}

\subsection{Multi-view Representation Consisency}
\label{sec:MV-SDS}

Although the aforementioned conditions can enhance the consistency of the generation status, considering the stochastic nature of diffusion process, it is crucial to align the generated  representations by controlling the noise sampling distribution for each viewpoint.
Therefore, we propose utilizing cross-view priors to identify the optimal noise distribution for each perspective, ensuring that multi-view representations are well-aligned and remain clear. To this end, we develop a \textbf{M}ulti-\textbf{V}iew \textbf{N}oise \textbf{R}esampling \textbf{S}trategy (\textbf{MV-NRS}) within the SDS Loss. MV-NRS consists of two steps: \emph{anchor noise initialization} and \emph{anchor noise optimization}, as shown in \cref{fig:p-nvs} (b).

To identify these noise distributions, we begin by pinpointing a specific set of multi-view noise, denoted as {anchor noise} $\epsilon^\text{Ac}_{v_1:v_N}$, to ensure the generated images under these noises are initially more closely aligned. 
%
Subsequently, we perform  resampling based on this anchor noise set, facilitating an initial alignment of the generated distributions.
The resampled $\epsilon^\text{Rs}$, with a small variance $\sigma^2$,  follows the distribution:
\begin{equation}\label{eq:resample}
    \epsilon^\text{Rs} \sim \mathcal{N}(\sqrt{1-\sigma^2}\epsilon^\text{Ac}, \sigma^2 \textbf{I})
\end{equation}

We recognize that the output of the 2D diffusion model demonstrates both invariance to linear transformations and robustness to small-scale nonlinear transformations. Consequently, it exhibits a degree of invariance to small-range viewpoint changes, which can be considered as local linear transformations.
Therefor, by aligning the inputs according to the viewpoints, we can ensure that the outputs align as well. 
Since the input of the UNet consists of a combination of rendered image latents and noises, we only need to align the noises.  
To achieve this, we just lift the driven view noise into a point cloud and render it onto the target views:
\begin{equation}\label{eq:ac}
\begin{aligned}
   &\epsilon^\text{Ac}_{v_{i+1},0} = \mathcal{R}_{P_{v_i,0}}(\epsilon^\text{Ac}_{v_{i},0}, v_{i+1}) + \epsilon^\text{rand}\odot M^\text{void}_{v_{i+1}, 0} \\
   &\epsilon^\text{Ac}_{v_{0},0} \sim \mathcal{N}(0, \textbf{I}), \quad \epsilon^\text{rand} \sim \mathcal{N}(0, \textbf{I}),
\end{aligned}
\end{equation}
where $s = 0$ denotes the initial training iteration  and $M$ is a mask that indicates the locations of voids in the rendered noise. 
Compared to random noise initialization, this method uses cross-view priors to build noise, enabling the capture of some small-scale noise distributions that are almost impossible to obtain through pure random sampling, particularly when the multi-view generative distributions are far apart.

\vspace{2pt}

Next, since the initial representations may not be perfectly aligned, we utilize multi-view gradient consistency to gradually finetune the anchor noises during the SDS training. Specifically, we have designed a {Resampling Retention Strategy}:

In each training iteration $s$, we first resampled a noise $\epsilon^\text{Rs}_{v_i,s}$  according to \eqref{eq:resample}.
Then, we decide whether to keep the resampled noise for updating the anchor noise by utilizing the multi-view gradient consistency score. The key idea is compute the gradients obtained from both the resampled noise and the anchor noise, and then assess their similarity with the gradients from the driven viewpoint. By comparing these similarities, we can determine whether to retain the resampled  noise.

In detail, for \(\epsilon^\text{Rs}_{v_i,s}\), we first compute the loss between a rendered image $I^{R}_{v_i,s}$ and denoised image $I^{D}_{v_i,s}$ from \(\epsilon^\text{Rs}_{v_i,s}\) , then backpropagate it to get the gradient \(grad^{D}_{v_i,s}\): 
\begin{equation}\label{eq:loss}
    \mathcal{L}^{D}_{v_i,s} = \mathcal{L}_{I}(I^{R}_{v_i,s}, I^{D}_{v_i,s})
\end{equation}
\begin{equation}\label{eq:grad}
    grad^{D}_{v_i,s} = \mathcal{BP}_{\Psi_{s-1}}(\mathcal{L}^{D}_{v_i,s})
\end{equation}
The cosine similarity is used to evaluate the consistency between this gradient and the applied gradient of the driven view $grad_{v_{i-1},s}$:
\begin{equation}\label{eq:score}
    S^{D}_{v_i,s} = \frac{grad^{D}_{v_i,s} \cdot grad_{v_{i-1},s}}{\|grad^{D}_{v_i,s}\| \|grad_{v_{i-1},s}\|}  
\end{equation}
Similarly, for anchor noise \(\epsilon^\text{Ac}_{v_i,s-1}\), we direct use the $I_{v_i,s-1}$ of previous training iteration to compute the gradient \({grad}^{P}_{v_i,s}\) the corresponding score \(S^{P}_{v_i,s}\).

If $ S^{D}_{v_i,s} > S^{P}_{v_i,s} $, the resampled noise is superior over the anchor noise in terms of gradient consistency. Therefore, we update the anchor noise and treat \(I^{D}_{v_i,s}\) as the current target image, \({grad}^{D}_{v_i,s}\) as the current applied gradient. Conversely, we retain the anchor noise and target image, using the corresponding gradient instead:
\begin{equation}\label{eq:update ac}
    \begin{aligned}
        I_{v_i,s}, &\ \epsilon^\text{Ac}_{v_i,s}, {grad}_{v_i,s} = \\
        & \begin{cases} 
I^{D}_{v_i,s}, \ \epsilon^\text{Rs}_{v_i,s}, \ {grad}^{D}_{v_i,s}
& \text{if } S^{D}_{v_i,s} > S^{P}_{v_i,s} \\
I_{v_i,s-1}, \ \epsilon^\text{Ac}_{v_i,s-1}, \ {grad}^{P}_{v_i,s} 
& \text{otherwise}
\end{cases}
    \end{aligned}
\end{equation}

After calculations for all viewpoints, we aggregate the gradients across all views, denoted as $ {Grad}_{s} $. Finally, we update the 3D model in the current training iteration:
\begin{equation}  
\begin{aligned}  \label{eq:acc grad}
    {Grad}_{s} = \sum_v  {grad}_{v_i,s} \\
\end{aligned}  
\end{equation}
\begin{equation}  \label{eq:update psi}
{\Psi_{s}} =  {\Psi_{s-1}} + \alpha_s \cdot {Grad}_{s}
\end{equation} 
For more detailed MV-NRS process, refer to 
Sec. {\color{red} 7}
.

\section{Experiments}
\subsection{Implementation Details}
\begin{figure*}[ht]
    \centering
    \includegraphics[width=1.0\linewidth]{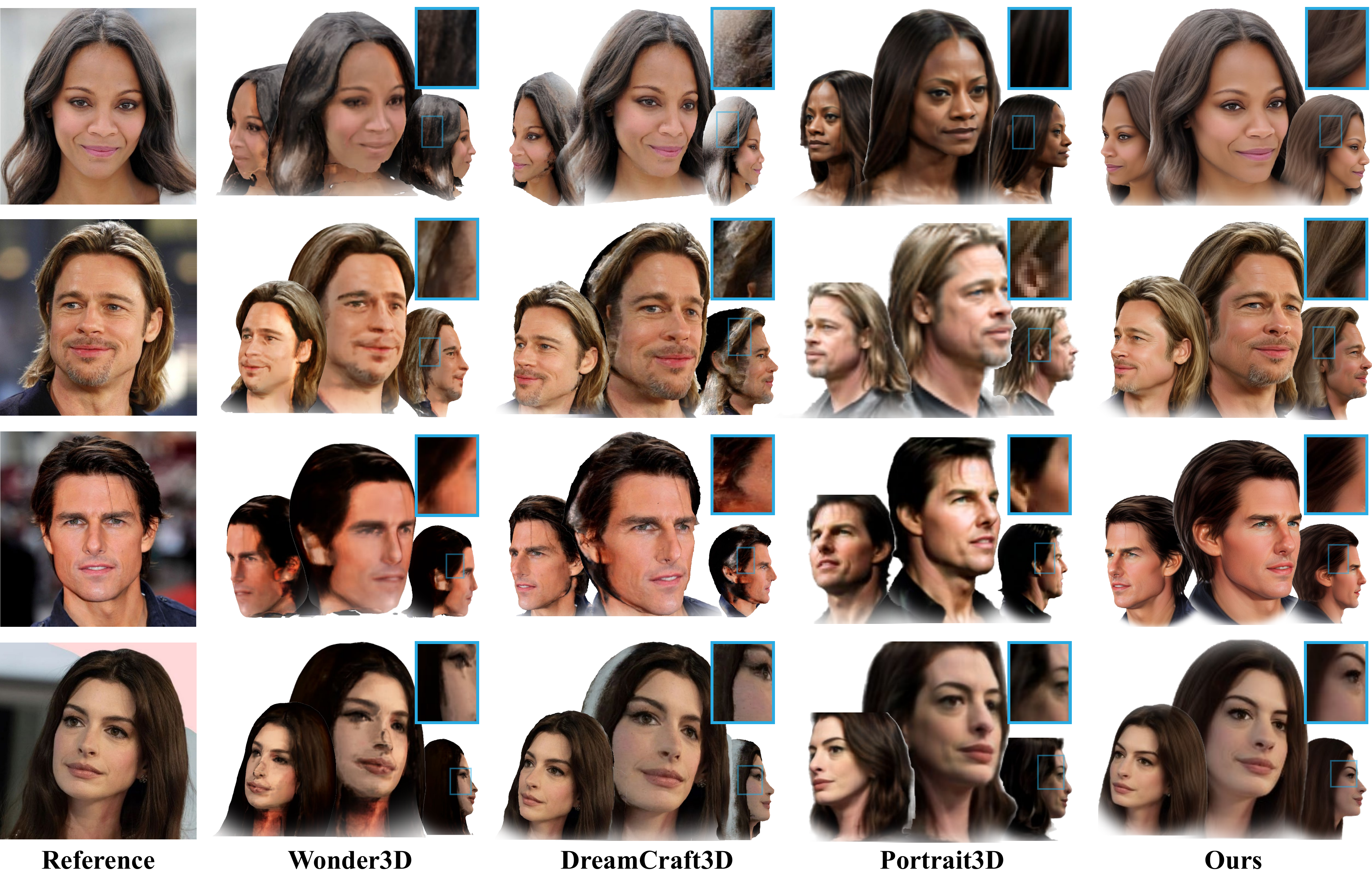}
    \caption{Qualitative comparison to SOTA approaches: Portrait3D~\cite{wu2024portrait3d}, Wonder3D~\cite{long2024wonder3d}, and DreamCraft3D~\cite{sun2023dreamcraft3d}. Our method presents the most photorealistic 3D portraits, with the most detailed textures in the face and hair strands. 
    \emph{\textbf{Zoom in for more detailed insights}.}
    }
    \label{fig:compare}
    \vspace{-2mm}
\end{figure*}

We employ triplane features with a resolution of 128x128 and a total of 96 channels, resulting in a rendered image resolution of 512x512. During the geometry restoration phase, we randomly sampling viewpoints across a full 360° azimuth and a pitch angle ranging from -30° to 30° for SDS training. In the texture restoration phase, we reconstruct images using 13 fixed viewpoints. The training of our HPDM uses synthetic dataset from GANs methods. 
In addition to these fixed perspectives, we randomly sample additional viewpoints and apply a multi-step denosing SDS loss (image-supervised) to improve rendering quality between the fixed viewpoints. All our experiments were conducted on a single A100 GPU. For more hyperparameters and training details, please refer to the 
Sec. {\color{red} 7}
.

\subsection{Qualitative Results}
We compare our results with those from open-sourced SOTA approaches: Portrait3D~\cite{wu2024portrait3d}, DreamCraft3D~\cite{sun2023dreamcraft3d}, and Wonder3D~\cite{long2024wonder3d}. Notably, Portrait3D is a text-to-3D method; in our case, we bypass the text-to-image step by directly providing the reference image.

From \cref{fig:compare}, it can be observed that the models generated by Wonder3D exhibit excessive smoothness and show a distinctly toy-like texture. While DreamCraft3D demonstrates some improvements, it completely loses reasonable texture in the profile view. Compared to the previous methods, Portrait3D can produce visually appealing models; however, it experiences identity variation issues and exhibits overly blurred textures and some artifacts in regions with complex hair patterns (e.g. the hair ends in the second row). In contrast, our model demonstrates the strongest fidelity and texture quality. We achieve an identity that is highly consistent with the reference and generating hair texture style that align appropriately with the reference. Furthermore, we can distinctly showcase the texture of nearly every strand of hair. Therefore, our propsed Portrait Diffusion surpasses SOTA methods and achieves the most detailed textures in 3D portraits.

\subsection{Quantitative Evaluation}
To comprehensively evaluate our generated head models, we employed three complementary metrics: CLIP-I for overall structural consistency, LPIPS for perceptual similarity, and ID metric for identity preservation. 
For the CLIP-I metric, we compute the cosine similarity of the CLIP features between images.
The evaluation is implemented by rendering images from five distinct viewpoints: front, left-frontal, left, right-frontal, and right. All five rendered images are included in the calculations for the three metrics alongside the reference image.

The \cref{tab:compare} shows our quantitative evaluation results. Our method achieves the highest CLIP-I score, indicating that our 3D portrait exhibits the greatest semantic consistency across various viewpoints. Additionally, we obtain the lowest LPIPS loss, demonstrating that our portraits maintain the most visually consistent appearance from different views. Furthermore, our approach achieves the highest ID score, confirming that our portraits exhibit the strongest identity consistency across multiple perspectives. The highest quantitative metrics indicate that our approach surpasses the generative quality of SOTA methods.

\begin{table}[t]
    \centering
    \begin{tabular}{lccc}
    \toprule
        Method & CLIP-I $\uparrow$ & LPIPS $\downarrow$& ID $\uparrow$\\
    \midrule
        Portrait3D &  0.9956&  0.4258& 0.1899\\
        Wonder3D &  0.9943&  0.4377& 0.3057\\
        DreamCraft3D &  0.9969&  0.4064& 0.2314\\
        Portrait Diffusion &  \textbf{0.9986}&  \textbf{0.3616}& \textbf{0.3440}\\
    \bottomrule
    \end{tabular}
    \caption{Quantitative comparison to SOTA approaches.}
    \label{tab:compare}
    \vspace{-4mm}
\end{table}

\subsection{Ablation Study}
\begin{figure}[t]
    \centering
\includegraphics[width=1.0\linewidth]{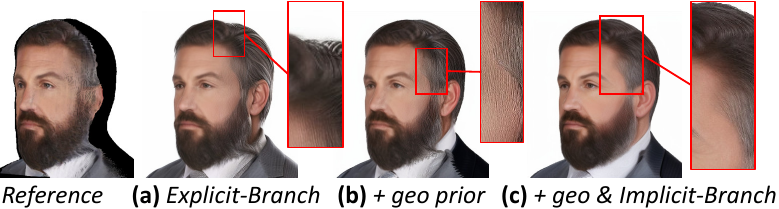}
    \caption{Visual Results for Ablation study on Hybrid Priors Diffusion Model. }
    \label{fig:abs diffusion}
    \vspace{-4mm}
\end{figure}

\textbf{Effectiveness of the Hybrid Priors Diffusion Model.}
The \cref{fig:abs diffusion} illustrates the visual results of the ablation study conducted to evaluate the effectiveness of the Hybrid Priors Diffusion Model.
Panel (a) displays results of conditioning only with Explicit-Branch, which aligns corresponding areas in the {reference} but reveals numerous deficiencies in both structure and texture. Panel (b) demonstrates the results achieved by incorporating geometric priors into the model. This addition fosters a more harmonious and unified geometry, significantly improving the overall shape fidelity of the 3D portrait. However, there are still texture striped artifacts present. Panel (c) further incorporates texture priors (Implicit-Branch), effectively eliminating all striped texture artifacts and showcasing exquisite details. This validates the effectiveness of our approach in leveraging multiple types of priors.

\begin{figure}[t]
    \centering
    \includegraphics[width=1.0\linewidth]{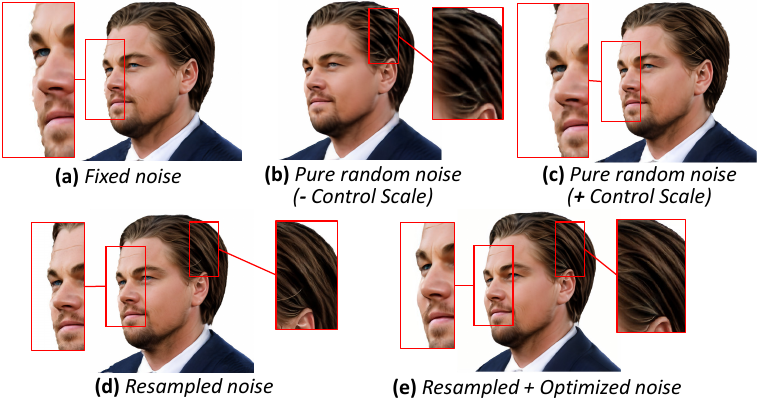}
    \caption{Visual Results for Ablation study on Multi-View Noise Resampling Strategy.}
    \label{fig:abs loss}
    \vspace{-4mm}
\end{figure}

\noindent\textbf{Effectiveness of the Multi-View Noise Resampling Strategy}.
The visual results displayed in \cref{fig:abs loss} illustrate the ablation study conducted to assess the effectiveness of the Multi-View Noise Resampling Strategy (MV-NRS).
Panel (a) displays the results with fixed noise during SDS loss, which lack randomness and reveal artifacts resulting from misalignment. Panel (b) and (c) show the outcomes with completely random noise: Panel (b) utilizes a lower control scale, resulting in a smoother generation representation while maintaining consistency, while Panel (c) applies a higher control scale, overly constraining the generation distribution and resulting in issues similar to Panel (a), though to a lesser degree. Panel (d) presents the results of MV-NRS without anchor noise optimization, resembling the outcomes in Panel (c). This demonstrates that without continuously updating anchor noise, alignment cannot be effectively improved.
Panel (e) presents our complete MV-NRS, which avoids both excessive smoothing and artifacts.
These results demonstrate that our MV-NRS, through both resampling and optimization, enables the fine-grained generation of detailed textures using HPDM. 

\section{Conclusion}
We introduced a Portrait Diffusion pipeline that generates detail-rich 3D full portraits. This pipeline consists of three main modules, including a GAN-prior Initialazation module, a Portrait Geometric Restoration module and a Mult-view Diffusion Refinement Module. Our Mult-view Diffusion Refinement Module incorporates a Hybrid Priors Diffusion model that effectively leverage multi-view priors for consistent status, and a Multi-View Noise Resampling Strategy to ensure consistent representations during the optimization. Qualitative and Quantitative assessments have shown that portraits produced by our pipeline exhibit superior detail and realism compared to state-of-the-art alternatives. Our Portrait Diffusion pipeline sets a new standard in 3D portrait generation, offering unparalleled texture detail and fidelity, and paving the way for future developments in computer vision, graphics, and digital art.

{
    \small
    \bibliographystyle{ieeenat_fullname}
    \bibliography{main}
}


\end{document}